# Occlusion-Aware Self-Supervised Monocular Depth Estimation for Weak-Texture Endoscopic Images

Zebo Huang[1] and Yinghui Wang[2]

[1] Jiangnan University, Wuxi 214122, China
[2] Engineering Research Center of Intelligent Technology for Healthcare, Ministry of Education, Wuxi 214122, China

**Abstract.** We propose a self-supervised monocular depth estimation network tailored for endoscopic scenes, aiming to infer depth within the gastrointestinal (GI) tract from monocular images. Existing methods, though accurate, typically assume consistent illumination, which is often violated due to dynamic lighting and occlusions caused by GI motility. These variations lead to incorrect geometric interpretations and unreliable self-supervised signals, degrading depth reconstruction quality. To address this, we introduce an occlusion-aware self-supervised framework. First, we incorporate an occlusion mask for data augmentation, generating pseudo-labels by simulating viewpoint-dependent occlusion scenarios. This enhances the model's ability to learn robust depth features under partial visibility. Second, we leverage semantic segmentation guided by non-negative matrix factorization (NMF), clustering convolutional activations to generate pseudo-labels in texture-deprived regions, thereby improving segmentation accuracy and mitigating information loss from lighting changes. Experimental results on the SCARED dataset show that our method achieves state-of-the-art performance in self-supervised depth estimation. Additionally, evaluations on the Endo-SLAM and SERV-CT datasets demonstrate strong generalization across diverse endoscopic environments.

**Keywords:** Endoscopy, Self-Supervised Learning, Monocular Depth Estimation, Semantic Segmentation.

## 1 Introduction

Endoscopy is widely used for diagnosing gastrointestinal diseases due to its non-invasive, painless nature and lack of cross-infection [1]. Depth estimation, a key technology for 3D gastrointestinal reconstruction, provides depth information from endoscopic images, enabling high-precision 3D models for tasks such as lesion identification, localization, and navigation [2]. However, depth estimation remains challenging due to occlusion from gastrointestinal folds, peristalsis, and lighting variability affecting texture and morphology [3].

Traditional monocular depth estimation methods, like Structure-from-Motion [4], SLAM [5], and Shape-from-Shading [6], rely on pixel correspondences and camera parameters, but struggle with lighting and occlusion. Deep learning-based methods, benefiting from stronger adaptability, are increasingly used for monocular depth estimation in endoscopy [7]. These methods often employ self-supervised learning, using information from consecutive frames or generating virtual endoscopic images to simulate real environments and reduce reliance on real depth data [8]. However, they often overlook occlusion and lighting-induced texture loss, reducing 3D reconstruction quality [9].

Gastrointestinal peristalsis and deformation exacerbate occlusion challenges. Some methods address this by using geometric priors or optical flow, but extreme occlusion scenarios still lead to errors [10,11]. Inspired by methods for outdoor scenes, interpretable masks could be applied to endoscopy to handle occlusions [12]. Lighting variations within the gastrointestinal tract complicate depth estimation, especially in low-light conditions. Semantic information, which shows geometric consistency, can improve depth estimation by helping the model focus on important regions and enhancing its robustness to lighting changes. Recent work has introduced semantic-aware information to preserve image details in low-light environments.

We propose two methods for generating pseudo-labels to address occlusions caused by peristalsis and texture loss from lighting variations, improving the model's understanding of gastrointestinal 3D structures. Our approach emphasizes handling occlusions and learning semantic texture information under varying lighting, setting it apart from existing methods. The main contributions of this work are as follows:

(1) A pseudo-label-based data augmentation method is proposed to address the gastrointestinal surface occlusion problem. This method combines pseudo-label depth maps, generated by both depth estimation and pose estimation networks, which are then used to train the depth estimation network. The approach enhances the model's ability to perform accurate depth estimation on the gastrointestinal surface in occluded environments.

(2) To address the challenge of semantic understanding in weak-texture images of the gastrointestinal tract under varying lighting conditions, we propose a method based on network feature clustering. The novelty of this approach lies in the introduction of non-negative matrix factorization (NMF) to cluster the activation features of the convolutional layers. This not only captures the local structure of the image but also identifies the semantic consistency across different regions, thereby improving the accuracy of semantic segmentation in weak-texture areas of endoscopic images. Additionally, it mitigates the issue of weak-texture information loss caused by lighting variations.

(3) We conduct a comprehensive ablation and comparative study to evaluate the effectiveness of the proposed method. The experimental results show that the proposed method achieves state-of-the-art performance and demonstrates effective generalization capabilities among self-supervised approaches.

## 2     Method

### 2.1     Overview

This paper focuses on addressing the issues of occlusion and lighting variations and proposes a self-supervised monocular depth estimation model, M-DASS (MASK Data

Augmentation and Semantic Segmentation). M-DASS is built upon AF-SfMLearner [11], which introduced appearance flow to handle severe inter-frame brightness fluctuations in endoscopic scenes. Appearance flow accounts for any changes in brightness patterns and induces generalized dynamic image constraints, thereby relaxing the assumption of photometric consistency. While AF-SfMLearner opens a new direction, it has limitations, particularly in addressing occlusion caused by gastrointestinal peristalsis and the loss of weak-texture information in the gastrointestinal tract due to lighting variations. Inspired by MASK data augmentation [13] and semantic segmentation pseudo-labeling methods [14], this paper builds upon AF-SfMLearner as the base framework and introduces corresponding MASK data augmentation and semantic segmentation pseudo-label methods. MASK data augmentation improves the model's learning capability by simulating occlusion scenarios from different viewpoints, while the semantic segmentation pseudo-label method utilizes complex semantic information from gastrointestinal images to guide more reliable supervision signals. The architecture of M-DASS is shown in Figure 1 and consists of six modules: appearance estimation, optical flow estimation, pose estimation, data augmentation, depth estimation, and semantic segmentation.

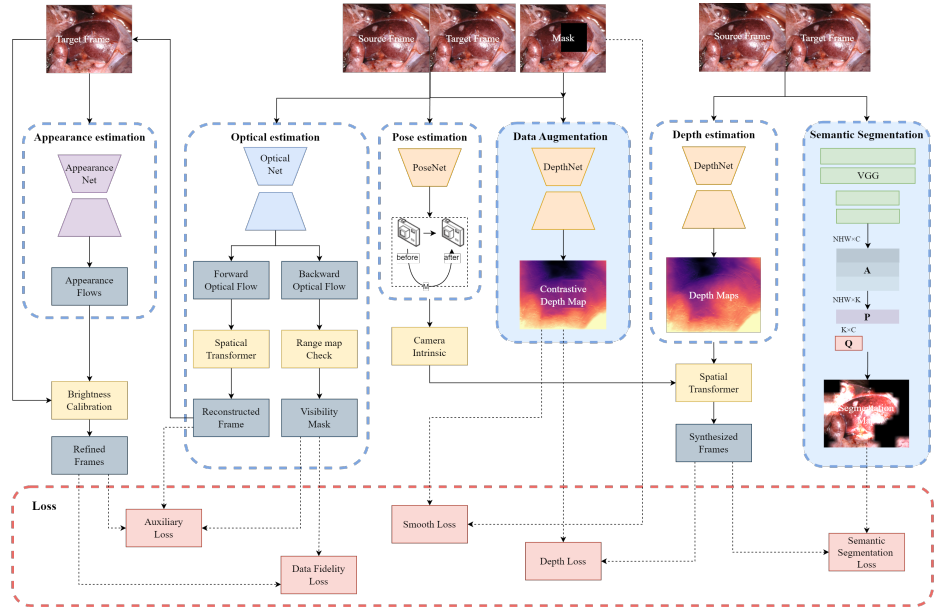

**Fig. 1.** M-DASS Self-supervised Monocular Image Depth Estimation Model

## 2.2 Data Augmentation

### Network Architecture
In the AF-SfMLearner [11], the pose estimation module acts as a collaborative component with the depth estimation module, providing the pose information of the endoscope as a reference for depth estimation, thereby enabling joint optimization. This is a common approach in self-supervised monocular depth estimation and ego-motion learning models. Building upon this, we retained the original pose estimation module unchanged and made improvements to the depth estimation module by adding a data augmentation module. In the depth estimation module, we introduced an occlusion Mask for data augmentation and generated pseudo-label depth maps. These pseudo-label depth maps are used to train the depth estimation network, enabling it to perform better in occluded environments.

The specific process is shown in the third and fourth dashed boxes from left to right in Figure 1. A Mask is added to the target frame image $I^t(p)$ with the width and height equal to one-quarter of the target frame's size. The target frame $I^t(p)$ with the added Mask is then concatenated with the source frame $I^s(p)$ and input into the DepthNet network to obtain the contrastive depth map. While generating the contrastive depth map, the target frame $I^t(p)$ and the source frame $I^s(p)$ are concatenated and fed into the PoseNet network to obtain the ego-motion pose transformation matrix. Once the ego-motion is estimated, the source frame depth can be projected onto the target frame's image plane. During training, the supervision signal for depth estimation comes from the distorted view synthesis obtained via differentiable bilinear sampling. Given two frames, the target frame $I^t(p)$ and the source frame $I^s(p)$, view synthesis is represented as shown in Equation (1).

$$h(p^{s \to t}) = [K \mid 0]M^{t \to s} \begin{bmatrix} D^t K^{-1} h(p^t) \\ 1 \end{bmatrix} \tag{1}$$

In this context, $h(p^{s \to t})$ and $h(p^t)$ represent the homogeneous pixel coordinates of the source and target frames, respectively. $K$ denotes the camera intrinsic parameters, $D^t$ represents the depth map of the target frame, and $M^{t \to s}$ is the ego-motion pose transformation matrix from the target frame $I^t(p)$ to the source frame $I^s(p)$. The matrix $\begin{bmatrix} D^t K^{-1} h(p^t), 1 \end{bmatrix}^T$ represents the transformation of pixel coordinates in the target frame $I^t(p)$ into 3D coordinates in the camera coordinate system. Here, $K^{-1}$ maps pixel coordinates to normalized camera coordinates. The normalized camera coordinates contain only directional information, without depth information. By multiplying these normalized coordinates by the depth map $D^t$, the scale-less directional vectors are extended to actual 3D spatial coordinates. The entire expression in Equation (1) represents the process of mapping points from the target frame image back to the source frame image, after incorporating depth information and pose transformation.

## Loss Function

In the depth estimation branch, the loss function for depth estimation consists of depth loss $L_{depth}$ and smoothness loss $L_{smooth}$, both of which are highly sensitive to variations in depth direction. The specific loss calculation method involves applying a spatial transformation to the source frame depth map $D_s$, resulting in the transformed depth map $D_s'$. This is then compared with the target frame depth map $D_t$, which has been processed using a Mask. The depth loss $L_{depth}$ is calculated using $L_1$ loss, as shown in Equation (2).

$$L_{depth} = \sum \left\| (D_t - D'_s) \odot M_t \right\|$$  (2)

where $M_t$ represents the portion of the mask that is not occluded.

We use smoothness loss to enforce the smoothness of depth predictions while preserving the fine details of image edges. Since depth discontinuities often occur at image gradients, we weight this cost using the image depth gradient, encouraging local smoothness in the depth map. $L_1$ loss is applied to the depth gradient, and the smoothness loss $L_{smooth}$ is computed as shown in Equation (3).

$$L_{smooth} = \left| \nabla D(p) \right| * e^{-\nabla \left| I'(p) \right|}$$  (3)

where $\nabla D(p)$ represents the depth map gradient, and $\nabla \left| I'(p) \right|$ denotes the image gradient.

Equation (3) promotes smoothness by penalizing the gradients of the depth map, while using the image gradient as a weight to ensure that areas with depth discontinuities, such as object boundaries, are not overly smoothed. This is especially important in complex scenarios, such as endoscopic images, where there is less texture variation, and depth estimation is more prone to blurring and discontinuities.

### 2.3    Semantic segmentation

**Network Architecture**
As shown in the semantic segmentation module in Figure 1, the feature maps extracted from the target and source frames by the VGG network have dimensions (H,W,C). These feature maps are concatenated and reshaped into a matrix $V$ of size (NHW,C). Then, Non-negative Matrix Factorization (NMF) is applied using the multiplicative update rule. NMF is a matrix decomposition algorithm used in multivariate analysis and linear algebra. Its core idea is to decompose a non-negative matrix $V_{n \times m}$ into two matrices, $P_{n \times k}$ and $Q_{k \times m}$, which allows the main structure of the data to be represented in a lower-dimensional space. By retaining the non-negativity constraint during dimensionality reduction, NMF can effectively capture local features in endoscopic images, providing strong support for clustering tasks. NMF inherently possesses clustering properties, as it can automatically cluster the columns of the matrix $V$ [15]. Specifically, for the matrix $V = (v_1,...,v_n)$, when the matrix H satisfies the orthogonality constraint, the error function shown in Equation (4) is minimized, achieving an approximation $V \approx PQ$, which is equivalent to optimizing K-means clustering.

$$\| V - PQ \|_F, P \geq 0, Q \geq 0$$  (4)

Here, the subscript F represents the Frobenius norm, which is used to measure the difference between two matrices.

After Non-negative Matrix Factorization (NMF), the matrix $V$ is decomposed into a matrix $P$ of size (NHW,K) and a matrix $Q$ of size (K,C), where K represents the number of semantic clusters, or the NMF factor. To provide a comprehensive understanding, the matrices $P$, $Q$, and the clustering effect of NMF are briefly illustrated in Figure 2.

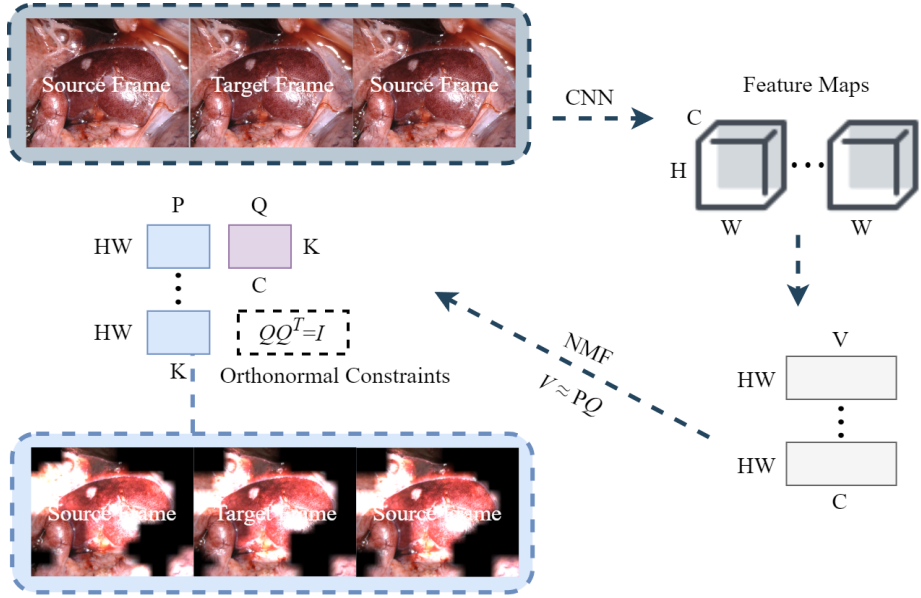

**Fig. 2.** The diagram illustrating the clustering effect of Non-negative Matrix Factorization

Due to the orthonormal constraint of Non-negative Matrix Factorization( $QQ^T = I$ ), each row of the matrix $Q \in {}^{(K,C)}$ can be viewed as a C-dimensional cluster center, corresponding to a coherent object between the views. The matrix $P$ is of size $(NHW, K)$, where its rows correspond to the spatial positions of all pixels from the N views. In general, matrix decomposition V≈PQ forces the product of each row of $P$ and each column of $Q$ to best approximate the C-dimensional features of each pixel in $V$. As shown in Figure 2, from the feature embeddings of all pixels in $V$, K(where K=4) semantic objects are clustered in the matrix $Q$, and the matrix $P$ contains the similarity between all pixels and the K semantic clusters. Therefore, $V$ can be further reshaped into N matrices of dimension $(NHW, K)$, which are then fed into a Softmax layer to construct the semantic segmentation map S, serving as a pseudo-label for our method.

**Loss Function**

Once the semantic segmentation pseudo-labels are obtained, a self-supervised constraint based on semantic consistency is designed, extending the photometric consistency between different views to the segmentation map. Given the predicted depth value $D(p_j)$ and the j-th pixel in the image, the corresponding position $p'_j$ in the source frame view can be calculated from the pixel $p_j$ in the target frame view using Equation (1). Then, through bilinear sampling, the distorted segmentation map $S'_i$ of the i-th source frame view can be reconstructed.

Finally, the semantic consistency objective $\mathcal{L}_{SS}$ is computed by calculating the pixel-wise cross-entropy loss between the distorted segmentation map $S'_i$ and the ground truth label obtained by transforming the segmentation map $S_0$ from the target frame, as shown in Equation (5).

$$L_{SS} = -\frac{1}{\parallel M_i \parallel_1} \sum_{j=0}^{HW} \Big[ oh(S_0)log(S'_{-1,j}) + oh(S_0)log(S'_{1,j}) \Big] M_{i,j} \tag{5}$$

Here, $oh(S_0)$ represents the one-hot encoding of the pixel values in the target frame segmentation map, $S'_{-1,j}$ denotes the distorted segmentation map of the source view at the j-th pixel, and $M_{i,j}$ is a binary mask that indicates the valid pixels from the i-th view to the reference view j.

## 3 Experiments

### 3.1 Dataset

The experiment uses an open-source endoscopy dataset, which includes several images of the gastrointestinal tract and their corresponding ground truth data. The dataset is divided into training, validation, and testing sets based on functionality.

- The SCARED dataset. The SCARED dataset is collected from the abdominal dissection of fresh pig cadavers and contains 35 endoscopic video segments. It provides ground truth data for depth maps, point clouds, and ego-motion. In this dataset, the maximum depth range of the depth maps is scaled to 150 mm, covering the depth range of most regions of the gastrointestinal tract.
- The Endo-SLAM dataset. The Endo-SLAM dataset is collected from the gastrointestinal organs of pigs, and the ground truth acquisition process is relatively challenging. This dataset includes ground truth for the pose of certain organs, as well as depth and pose ground truth for synthetic images, making it suitable for evaluating the performance of models in complex environments.
- The SERV-CT dataset. The SERV-CT dataset is extracted from the pig torso cadaver and contains 16 pairs of stereo images, along with ground truth data for depth and disparity. In this dataset, the upper limit of the depth map is set to 180 mm, ensuring coverage of almost the entire valid depth range.

## 3.2    Evaluation Metrics

Similar to the work of AF-SfMLearner [11], we adopt standard evaluation metrics: Absolute Relative Error (Abs-Rel), Square Relative Error (Sq-Rel), Root Mean Squared Error (RMSE), Root Mean Squared Error in log space (RMSE-log), and accuracy metrics ($\delta$). In Table 1, $d$ and $d^*$ represent the predicted depth values and the corresponding ground truth values, respectively, while $D$ denotes a set of predicted depth values.

**Table 1.** Evaluation Metrics for Depth Estimation

| Evaluation Metrics | Formula |
| --- | --- |
| Abs-Rel | $\dfrac{1}{|D|}\sum_{d\in D}\left|d^*-d\right|/d^*$ |
| Sq-Rel | $\dfrac{1}{|D|}\sum_{d\in D}\left|d^*-d\right|^2/d^*$ |
| RMSE | $\sqrt{\dfrac{1}{|D|}\sum_{d\in D}\left|d^*-d\right|^2/d^*}$ |
| RMSE-log | $\sqrt{\dfrac{1}{|D|}\sum_{d\in D}\left|\log d^*-\log d\right|^2}$ |
| $\delta$ | $\dfrac{1}{|D|}\left|\left\{d\in D\left|\max\left(\dfrac{d^*}{d},\dfrac{d}{d^*}<1.25\right)\right.\right\}\right|\times100\%$ |

## 3.3    Ablation Study

We conducted an ablation study (Table 2) to evaluate the contributions of the MASK-based data augmentation and semantic segmentation modules to depth estimation performance. The baseline model (ID-1) exhibited poor performance, underscoring the challenge of estimating depth in complex endoscopic scenes without additional enhancements. Introducing MASK data augmentation (ID-2) led to substantial improvements across all evaluation metrics, demonstrating its effectiveness in addressing occlusions. However, extending the training to 30 epochs (ID-3) resulted in performance degradation, suggesting that excessive training may cause overfitting and reduce generalization. Incorporating the semantic segmentation module alone (ID-4) improved both the square relative error and root mean square error, indicating its positive impact on depth estimation in weakly textured regions. When combining MASK with semantic pseudo-labeling guided by non-negative matrix factorization (NMF) with a factor of 3 (ID-5), accuracy improved further, highlighting the importance of enforcing semantic consistency. Optimizing the NMF factor to 4 (ID-6) yielded the best overall performance, suggesting that finer semantic granularity enables more precise modeling of local structures. Across all experiments, semantic consistency loss, smoothness loss, and depth loss were applied to enhance training stability. Depth map visualizations (Figure 3) further illustrate the progressive improvements. In summary, MASK data augmentation significantly reduces depth estimation errors in occluded regions, while semantic segmentation with optimized NMF enhances performance in weakly textured areas. The combination of both modules, as demonstrated by ID-6, achieves superior robustness and generalization in challenging endoscopic environments.

**Table 2.** Ablation experiment, where DA represents the data augmentation module and SS represents the semantic segmentation module.

| ID | epoch | DA | SS | Abs-Rel↓ | Sq-Rel↓ | RMSE↓ | RMSE-log↓ | δ ↑ |
|----|-------|----|----|----------|---------|-------|-----------|-----|
| 1 | 20 | | | 0.072 | 0.696 | 6.264 | 0.103 | 0.946 |
| 2 | 20 | √ | | 0.065 | 0.572 | 5.730 | 0.092 | 0.962 |
| 3 | 30 | √ | | 0.067 | 0.587 | 5.789 | 0.094 | 0.954 |
| 4 | 20 | | √ | 0.064 | 0.544 | 5.556 | 0.090 | 0.963 |
| 5 | 20 | √ | √ | 0.063 | 0.537 | 5.544 | 0.089 | 0.966 |
| 6 | 20 | √ | √ | 0.061 | 0.526 | 5.499 | 0.087 | 0.968 |

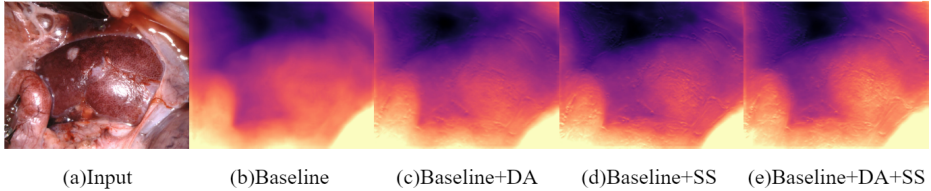

(a)Input          (b)Baseline     (c)Baseline+DA     (d)Baseline+SS     (e)Baseline+DA+SS

**Fig. 3.** Ablation experiment visualization

### 3.4 Comparison Experiments

We compared our proposed model with several representative monocular depth estimation methods, including classic models for outdoor scenes—SfMLearner [16], DeFeat-Net [17], and Monodepth2 [18]—as well as the state-of-the-art endoscopic model AF-SfMLearner. We tested these models on the SCARED and SERV-CT datasets and compared the results. Table 3 shows a comparison of the evaluation metrics for each model tested on the SCARED dataset. Our model significantly outperforms all the compared models. Notably, the Square Relative Error which is highly sensitive to depth errors, shows a marked improvement. Due to factors such as lighting variations, non-Lambertian reflections, and severe brightness fluctuations, models often fail to capture the weak texture information on the inner wall of the digestive tract, which typically leads to unstable supervision signals. Our model, however, is able to effectively handle intense brightness fluctuations and yields superior results.

**Table 3.** Model Evaluation Metrics on the SCARED Dataset

| Method | Abs-Rel↓ | Sq-Rel↓ | RMSE↓ | RMSE-log↓ | δ ↑ |
|--------|----------|---------|-------|-----------|-----|
| SfMLearner[16] | 0.079 | 0.879 | 6.896 | 0.110 | 0.947 |
| DeFeat-Net[17] | 0.077 | 0.792 | 6.688 | 0.108 | 0.941 |
| Monodepth2[18] | 0.071 | 0.590 | 5.606 | 0.094 | 0.953 |
| AF-SfMLearner[11] | 0.072 | 0.614 | 5.891 | 0.099 | 0.949 |
| Ours | **0.061** | **0.526** | **5.499** | **0.087** | **0.968** |

Table 4 shows the comparison experiment results of the models tested on the SERV-CT dataset. According to the evaluation metrics, our method demonstrates significant improvement over AF-SfMLearner. Notably, DeFeat-Net outperforms other methods in terms of Absolute Relative Error and Accuracy. This is because the SERV-CT dataset is relatively small, and DeFeat-Net incorporates self-supervised representation learning and multi-task learning, which enables the model to more effectively extract information from small datasets. However, during training, the model may overfit, leading to higher accuracy, which suggests that while the model may perform well on the SERV-CT dataset, its performance on larger datasets could be less effective.

**Table 4.** Comparison of Model Evaluation Metrics on the SERV-CT Dataset

| Method | Abs-Rel↓ | Sq-Rel↓ | RMSE↓ | RMSE-log↓ | $\delta$ ↑ |
|---|---|---|---|---|---|
| SfMLearner[16] | 0.151 | 3.917 | 17.451 | 0.191 | 0.779 |
| DeFeat-Net[17] | 0.114 | 3.946 | 12.588 | 0.153 | 0.873 |
| Monodepth2[18] | 0.118 | 3.532 | 13.681 | 0.149 | 0.829 |
| AF-SfMLearner[11] | 0.123 | 2.343 | 13.846 | 0.156 | 0.837 |
| Ours | 0.117 | 2.219 | 13.301 | 0.148 | 0.853 |

In Figure 4, we provide a qualitative comparison. Compared to Monodepth2 and AF-SfMLearner, our model is able to recover the weak texture depth information of the gastrointestinal tract walls more effectively.

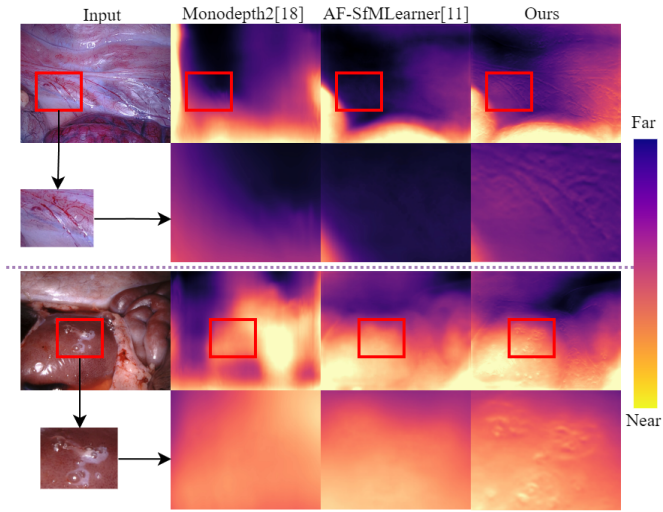

**Fig. 4.** Visualization comparison of depth estimation between Monodepth2, AF-SfMLearner and our model

Additionally, we directly validated the model trained on the SCARED dataset on the SERV-CT and Endo-SLAM datasets without any fine-tuning, only adjusting the resolution of the test frames to 320×256 pixels. The results for different datasets are shown in Figure 5, where it is evident that our method demonstrates strong generalization capability.

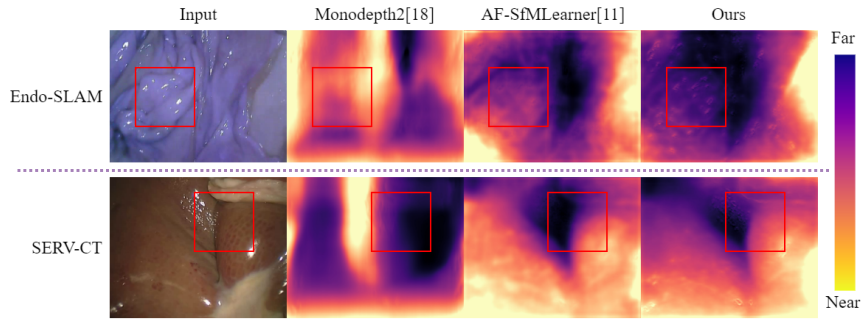

**Fig. 5.** Visualization of the comparison of generalization ability across different models

## 4    Conclusion

In this paper, we propose a novel self-supervised monocular depth estimation model, M-DASS, which significantly mitigates the issues of occlusion and semantic information loss commonly encountered in endoscopic self-supervised methods by combining MASK data augmentation with semantic segmentation pseudo-labeling for monocular depth estimation. Additionally, we design a new semantic segmentation loss, which, when combined with depth loss and smoothness loss, provides regularization constraints for the model. Compared to previous self-supervised monocular depth estimation models, our approach not only demonstrates higher immunity to occlusion, enabling more accurate 3D information for medical applications, but also integrates MASK data augmentation and semantic segmentation pseudo-labels to guide more reliable depth estimation supervision signals. This methodology exhibits broad applicability for self-supervised learning in complex visual tasks.

**Acknowledgments.** This work was supported by the National Natural Science Foundation of China (No. 62172190), in part of the "Double Creation" Plan of Jiangsu Province (Certificate: JSSCRC2021532), and in part of the "Taihu Talent-Innovative Leading Talent" Plan of Wuxi City (Certificate Date:202110).